\documentclass{article}

\PassOptionsToPackage{numbers}{natbib}
 \usepackage[main, final]{neurips_2026}
\usepackage[utf8]{inputenc} % allow utf-8 input
\usepackage[T1]{fontenc}    % use 8-bit T1 fonts
\usepackage{hyperref}       % hyperlinks
\usepackage{url}            % simple URL typesetting
\usepackage{booktabs}       % professional-quality tables
\usepackage{amsfonts}       % blackboard math symbols
\usepackage{nicefrac}       % compact symbols for 1/2, etc.
\usepackage{microtype}      % microtypography
\usepackage{xcolor}         % colors

\usepackage{graphicx}
\usepackage{amsmath, amssymb}
\usepackage{bbm}

\usepackage[ruled,vlined,linesnumbered]{algorithm2e}
\usepackage{wrapfig}
\usepackage{caption}

\usepackage{makecell}
\usepackage{multirow}

\title{Ego-HMB: Human Motion Bridging from Egocentric Images via A Motion Bridge Diffusion Model}

\author{%
  \parbox{\textwidth}{%
    \centering
    Gaoge Han$^{1,2}$ \quad
    Yongkang Cheng$^{1,2}$ \quad
    Shaoli Huang$^{2}$ \quad
    Jing Zhang$^{5}$ \\[0.5em]
    Mingming Gong$^{3,1}$\thanks{Co-corresponding authors:
    Mingming Gong and Tongliang Liu.
    Emails: \texttt{mingming.gong@unimelb.edu.au} and
    \texttt{tongliang.liu@sydney.edu.au}, respectively.} \quad
    Fakhri Karray$^{1}$ \quad
    Tongliang Liu$^{4,1}$\footnotemark[1] \\[0.8em]
    {\small
      $^{1}$Mohamed bin Zayed University of Artificial Intelligence \\[0.2em]
      $^{2}$AgiBot \quad
      $^{3}$The University of Melbourne \\[0.2em]
      $^{4}$The University of Sydney \quad
      $^{5}$Wuhan University
    } \\[0.6em]
    {\small\texttt{hangaoge@gmail.com}}
  }%
}

\begin{document}

\maketitle

\begin{abstract}
In this paper, we move beyond the conventional setting of motion in-betweening that relies on structured pose inputs, and instead propose a new paradigm that is conditioned on egocentric (first-person) visual observations, thereby reducing the requirements on both users and upstream systems; we term this task Egocentric Human Motion Bridging. This is motivated by realistic consumer scenarios in VR/AR and robotics, where users typically have access only to raw egocentric imagery (e.g., from wearable devices) without reliable human pose annotations or motion capture signals. Unlike conventional motion interpolation in human pose-conditioned settings or motion generation conditioned on a complete temporal language or image context, the egocentric perspective introduces large viewpoint variations, partial body visibility, and severe temporal ill-posedness due to cross-modal endpoint constraints. To this end, we propose Ego-HMB, a unified diffusion-based model that internalizes both endpoint human pose estimation capability and temporally consistent 3D motion infilling ability within a single model through two training stages. Specifically, we first introduce Egocentric-to-Canonical Pretraining (pretraining stage), where we design a hierarchical curriculum with progressively increasing temporal spans. Then, we present a novel Motion Bridge Diffusion Training with endpoint and physics awareness as a fine-tuning stage, where our bridge diffusion diffuse from neighboring poses to enhance temporal continuity while preserving the global constraints between the two endpoints of a 3D motion sequence. Extensive experiments and comparisons with one-stage baselines (a generative model conditioned directly on two egocentric images) and two-stage baselines (estimation models followed by interpolation models) demonstrate that our approach consistently outperforms these baselines.
Code is available at \url{https://github.com/gaogehan/Ego-HMB}.
\end{abstract}

\section{Introduction}

Human motion generation based on deep learning and diffusion models has achieved substantial success in VR/AR and robotics, significantly reducing the cost of motion creation for both professionals and general users while enabling diverse forms of interactive entertainment. Existing approaches can be broadly categorized into several classes: (1) multimodal-to-motion generation (typically conditioned on text or audio); (2) motion estimation (usually from video or monocular images); and (3) motion in-betweening/interpolation (given starting and ending poses). These tasks generally require dense user inputs or fully specified 3D pose sequences.

However, in common VR/AR and robotic application scenarios, users who intend to obtain motion sequences often have access only to sparse image representations, rather than complete pose trajectories. Generating full motion sequences from such sparse image inputs is therefore practically significant, as it reduces both user expression cost and storage overhead (by relying on sparse images instead of dense motion data).

To address this realistic requirement, we propose a new task termed egocentric human motion bridging task. In essence, the objective is to generate a coherent motion segment from two endpoint egocentric images. This task differs from existing paradigms in several aspects: 1) Unlike other cross-modal generation tasks, such as text- or scene-to-motion generation, where the conditioning signal usually describes a holistic description of the entire motion to be generated, our task is constrained by implicit start and end conditions. 2) It is not standard motion in-betweening/interpolation, since users are not required to provide complete 3D starting and ending poses. The notion of bridging refers to leveraging image modalities distinct from motion representations to connect human motion endpoints and synthesize the intermediate motion segment.

\begin{figure}[t]
  \centering
  \includegraphics[width=\linewidth]{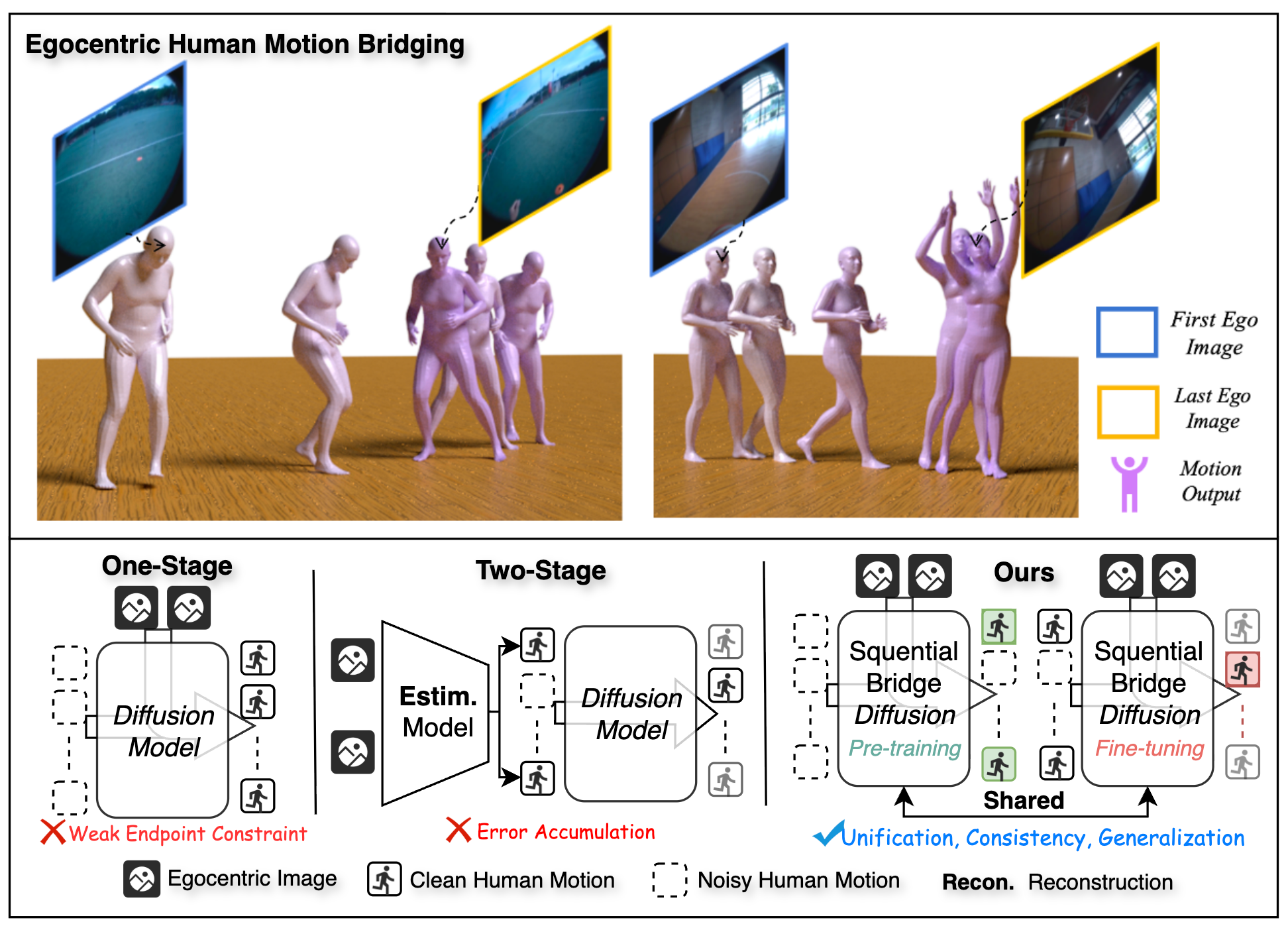}
  \caption{Egocentric Human Motion Bridging and Methods Comparison. Top: Egocentric human motion bridging from first-person observations to produce coherent intermediate motions; Bottom: Comparison of one-stage, two-stage, and Brownian bridge diffusion pipelines with improved training consistency and reduced error accumulation.
}
  \label{fig:ego-pipline}
\end{figure}

Therefore, how to effectively address the motion bridging task remains insufficiently and unsystematically explored in the existing literature. Based on current technical paradigms, two intuitive solution paths can be derived, both relying on the adaptation or composition of existing model frameworks, and thus serving as strong baselines for comparison with our method, as shown in Fig.~\ref{fig:ego-pipline}.

The first approach treats sparse ego images as conditioning signals and trains a conditional generative model directly. For example, diffusion-based text/scene-to-motion models such as~\cite{tevet2022mdm,Patel_2025_ICCV} can be adapted by replacing text or audio inputs with ego-image features. However, this strategy requires large-scale paired data, is vulnerable to temporal inconsistency in high-dimensional sequence generation, and lacks explicit structural constraints to ensure accurate alignment with the desired start and end states. The second approach follows a two-stage pipeline: first estimates 3D endpoint poses from input images~\cite{yi2024egoallo}, and then uses a motion in-betweening model~\cite{qin2022motion,Patel_2025_ICCV} to generate intermediate motion. While modular and intuitive, this design suffers from error accumulation, as inaccuracies in endpoint estimation can propagate and cause trajectory drift. It also introduces higher structural complexity, model size, and inference cost than an end-to-end formulation.

Our core motivation for this task is twofold. First, sparse endpoint ego observations make the problem highly ill-posed, while self-occlusion and limited body visibility further increase endpoint pose uncertainty. Thus, strong pose priors are needed to constrain the solution space beyond explicit conditioning. Second, the generated intermediate motion must jointly satisfy endpoint consistency, temporal continuity and plausibility, avoiding discontinuities or unstable body structures.

To address these challenges, we propose a unified bridge diffusion-based generative model trained in two stages, enabling a single framework to internalize both endpoint human pose estimation and physically consistent 3D motion generation, as presented in Fig.~\ref{fig:ego-pipline}. Our method has two key features: 1) we propose egocentric-to-canonical endpoint pretraining with a time-span scheduling strategy, which strengthens the model’s endpoint pose priors and enables more accurate endpoint fitting and alignment; and 2) we introduce the first motion bridge diffusion model that diffuses from neighboring motion frames. This design preserves end-to-end structural consistency while promoting physically continuous and plausible motion generation. Overall, our method avoids the strong coupling difficulty and error accumulation inherent in prior approaches.

% Specifically, our method performs pre-training and fine-tuning within a single generative model: 1) Egocentric-to-Canonical Pretraining.
% We design a hierarchical curriculum with progressively increasing temporal span, enabling the model to learn an accurate mapping from ego images to 3D poses. The curriculum follows a multi-stage difficulty scheduling strategy. 2) Bridge Diffusion Model Fine-tuning with Endpoint and Physics Awareness. We further fine-tune the model to generate intermediate frames while accurately modeling global constraints imposed by the two endpoints and maintaining physical temporal continuity. 

To rigorously validate the effectiveness of our approach, we implement the two aforementioned strategies as strong baselines and conduct systematic ablation studies on all major components of our framework. Experiments are performed on the  EE4D-Motion dataset~\cite{Patel_2025_ICCV}, a large-scale motion capture benchmark comprising 208 hours of high-quality mocap data. Extensive quantitative and qualitative evaluations consistently demonstrate that our method achieves significant improvements over competing approaches across multiple metrics.

Our contributions can be summarized as follows:

\begin{itemize}
    \item We introduce a new task, egocentric human motion bridging task, which aims to generate a coherent 3D motion segment from sparse egocentric image observations, addressing a practically relevant yet underexplored problem setting.
    \item We propose a unified bridge diffusion-based framework that jointly internalizes endpoint pose estimation and temporally continuous 3D motion completion through a two-stage training paradigm, enabling structurally consistent and physically plausible motion generation within a single model.
    \item We conduct comprehensive comparisons against strong baselines, including both direct conditional generation (one-stage) and estimation–interpolation pipelines (two-stage), and perform systematic ablation studies. Extensive quantitative and qualitative results demonstrate the clear improvements of our approach.
\end{itemize}

\section{Related Work}
% \subsection{}

\noindent \textbf{Human Motion Task:} Human Motion Modeling has gradually become a prominent research direction in computer vision and computer graphics. On the one hand, human motion exhibits strong nonlinearity, complex temporal dependencies, and multimodal characteristics, which pose significant challenges for modeling. On the other hand, it has broad application value in animation synthesis, virtual reality, human–computer interaction, robotic control, and behavior understanding.

Human motion tasks typically focus on generating or predicting plausible sequences of human joint movements under given conditions. The core challenge lies in modeling high-dimensional, structured, and strongly constrained spatiotemporal data. Depending on the task formulation, existing research can be broadly categorized into motion prediction~\cite{li2022skeleton,chen2023humanmac,xu2023auxiliary,tang2025stochastic}, motion generation~\cite{han2025reindiffuse,han2024hutumotion,lin2023motionx, zhang2024motiondiffuse,tevet2022human,petrovich2022temos, karunratanakul2023gmd}, motion interpolation~\cite{harvey2020robust,qin2022motion,cohan2024flexible,hwang2025scenemi}. Besides, motion generation tasks can be categorized according to the input modality, with common settings including text-to-motion \cite{han2024hutumotion,han2025reindiffuse,zhou2024emdm,tevet2022human}, audio-to-motion \cite{cheng2025conditional,cheng2025didiffges,cheng2025holegest}, and scene-to-motion \cite{wang2021sceneaware,Patel_2025_ICCV}.

% 根据输入的条件划分：常见的有text-to-motion, audio-to-motion, image-to-motion.

In this work, we focus on the growing interest in leveraging first-person perspectives to enhance perception for embodied intelligence and VR/AR applications, enabling the generation of human motion. To this end, we introduce a new task termed Egocentric Human Motion Bridging. Specifically, the goal is to directly generate intermediate motion sequences from sparse egocentric image observations.

% 

% Unlike prior motion prediction or interpolation settings that operate purely within the motion domain, our task formulates motion completion as a cross-modal problem, bridging visual inputs and structured human motion representations. This setting is motivated by practical user scenarios in virtual reality and robotics, where first-person visual data can be readily acquired, and there is a demand to directly recover temporally dense, full-body motion in the intermediate time span. By explicitly modeling this cross-modal bridging process, our task extends conventional human motion completion toward more realistic and application-driven settings.

% \subsection{Bridge Diffusion Model}
\noindent \textbf{Bridge Diffusion Models:}
The Bridge Diffusion Models (BDMs)~\cite{li2023bbdm,zhou2024denoising,lyu2024frame,lyu2025tlb} are a recently proposed variant of diffusion models that incorporates endpoint constraints into the generative process by leveraging the Brownian bridge stochastic process. Unlike standard denoising diffusion probabilistic models (DDPM)~\cite{ho2020denoising}, which learn to reverse a diffusion trajectory from pure Gaussian noise to the data distribution, BDM constructs a conditional stochastic path between two sources. This formulation explicitly enforces boundary consistency during generation.
% and is therefore particularly suitable for tasks that require strong conditional alignment or interpolation between given states

So far, Bridge Diffusion has primarily been explored in image transformation and interpolation tasks. Brownian Bridge Diffusion Model (BBDM)~\cite{li2023bbdm} introduces a brownian bridge diffusion framework for image-to-image translation. By treating the source and target images as the two endpoints of a stochastic bridge, the model learns a conditional diffusion trajectory that smoothly transforms one image domain into another. Denoising Diffusion Bridge Models (DDBMs)~\cite{zhou2024denoising} model distribution-to-distribution translation via diffusion bridges. By learning the bridge score and solving the associated stochastic differential equation, they offer a unified framework related to score-based diffusion and OT/flow matching, with improved results on image translation. TLB-VFI~\cite{lyu2025tlb} applies the Brownian bridge formulation in latent space for video frame interpolation. In this setting, the first and last frames serve as fixed endpoints, and intermediate frames are generated along a time-aware stochastic bridge.

In this work, we are the first to extend Bridge Diffusion to human motion modeling. We resolve the incompatibility of Bridge Diffusion with multi-frame sequence data. We formulate human motion bridging as a cross-modal stochastic bridge that connects sparse egocentric image observations to a complete motion sequence. Unlike image translation or completion, our task requires inferring full-body intermediate motion trajectories from sparse visual inputs under strong structural and temporal constraints. To this end, we design a tailored adaptation of the bridge diffusion framework to generate temporally coherent, kinematically plausible human motion while respecting observational guidance.

\section{Method}

% 我们将直观上使用estimation 和in-between模型去解决这个human motion bridging任务的形式内化到一个diffusion生成模型中，使用两个训练方式去在同一生成模型中内化两种能力。在这个章节，我们将详细介绍1. 时间跨度递进的预训练curriculum，这促进了我们使用更多的pair数据去scaling图像帧对应的pose estimation的能力。2. 首次提出motion bridging diffusion model 从邻近帧去扩散生成连续motion而不是纯噪声，有助于从端点pose恢复中间motion和建模平滑变化。
In this section, we first formulate our new egocentric human motion bridging task and then detail our technological innovations: 1) a short-to-long curriculum endpoint pretraining with progressively increasing temporal spans, which facilitates scaling the pose estimation capability from image-frame pairs by leveraging a larger amount of paired data; and
2) the first motion bridging diffusion model, which generates continuous motion by diffusing from neighboring frames rather than from pure noise, thereby enabling the recovery of intermediate motion from endpoint poses and improving the modeling of smooth temporal transitions.

%我们的方法的亮点是第一和第二个阶段都使用同一个diffusion进行学习训练。

\subsection{Problem Formulation}

% We denote the camera’s 6-DoF trajectory by $T_{1:N} = (T_1, T_2, \ldots, T_N)$. Such trajectories are readily available in real time on modern wearable platforms through inertial SLAM systems~\cite{15}.

Given an egocentric image sequence $I_{1:N} = (I_1, I_2, \ldots, I_N)$ acquired by a head-mounted camera, each frame satisfies $I_i \in \mathbb{R}^{H \times W \times 3}$. The corresponding 3D human motion is denoted by $X_{1:N} = (X_1, X_2, \ldots, X_N)$, where each pose $X_i$ is represented with SMPL-X~\cite{pavlakos2019smplx} parameters as $X_i = (R_i^r, t_i^r, \theta_i, \beta_i)$. Here, $R_i^r \in \mathbb{R}^3$ and $t_i^r \in \mathbb{R}^3$ denote the global root rotation and translation, $\theta_i \in \mathbb{R}^{21 \times 3}$ contains the local joint rotations, and $\beta_i \in \mathbb{R}^{10}$ captures the body shape. Following standard practice, we treat body shape as time-invariant, so $\beta_i = \beta_j$ for all $i, j$. 

%, T_s, T_e , and $T_s$ and $T_e$ represent the corresponding head-mounted camera poses  and the associated camera states
\textbf{Egocentric Human Motion Bridging Task} aims to generate a plausible motion segment that connects two temporally sparse egocentric observations. Formally, the task is to model $p(X_{1:N} \mid I_s, I_e)$, where $I_s$ and $I_e$ denote the egocentric images captured at the start and end of the target interval. The output is a motion sequence $X_{1:N}$ that bridges the two observations, such that the generated motion is consistent with both endpoint images. The term \emph{bridging} highlights precisely this property: the model must rely on egocentric visual cues, instead of explicit motion parameters, to bridge two motion endpoints and synthesize the missing motion segment. This task is particularly challenging because the input is highly ill-posed. Egocentric images only provide partial first-person observations of the surrounding scene, while the wearer’s body is largely invisible. As a result, the model must infer the latent human pose and dynamics from limited visual evidence, while ensuring temporal consistency with both endpoint observations.

% Compared with fully cross-modal motion generation tasks, such as text-to-motion or audio-to-motion, this problem is not purely open-ended generation, since the motion is constrained by both the starting and ending conditions. On the other hand, it also differs from classical motion in-betweening or interpolation, where complete 3D poses at the two endpoints are typically given as input. In our setting, the endpoint conditions are provided in the image modality rather than directly in the motion space, and the goal is to use these visual observations to infer and connect the underlying motion trajectory. 

\subsection{Egocentric-to-Canonical Endpoint Pretraining}

%更多内容 In the pretraining stage, we aim to learn a reliable mapping from egocentric observations to canonical 3D human poses at the first and last frames, which cast learning as an egocentric motion estimation problem. By decomposing each motion sequence into multiple paired endpoint samples with different temporal gaps, we significantly enlarge the amount of paired training data. This richer supervision helps the model learn a more reliable mapping from egocentric observations to canonical 3D poses, and alleviates the ill-posedness of egocentric human motion estimation caused by limited visibility, self-occlusion, and depth ambiguity.

%简洁版本
In the pretraining stage, we learn a reliable mapping from egocentric observations at the first and last frames to canonical 3D human poses, treating the task as egocentric motion estimation. By decomposing each motion sequence into multiple endpoint pairs with different temporal gaps, we increase the paired training data and reduce the ambiguity caused by limited visibility, self-occlusion, and depth uncertainty.

%, and camera trajectory conditioning $(T_s, T_e)$ and $(\tilde{T}_{s}, \tilde{T}_{e})$ , \tilde{T}_{(s,e)}
We pretrain the model with a conditional bridge diffusion objective on endpoint pose estimation. Specifically, given a motion window $X^{1:N}$, egocentric image features $(O_s, O_e)$, we construct a pretraining sample by selecting a pair of subframes $(X^s, X^e)$ with temporal gap $g=e-s$, where $g$ is progressively scheduled from short to long values. The selected endpoint frames are placed at the two ends of a fixed-length window, yielding $(\tilde{X}_{s}, \tilde{X}_{e})$. Only the corresponding conditioning inputs $(\tilde{O}_{s}, \tilde{O}_{e})$ remain visible at the two endpoint positions, while all intermediate frames are masked out. The model is then trained to learn $p_\theta(\tilde{X}_{(s,e)}\mid \tilde{O}_{(s,e)})$ under the same bridge diffusion framework. In our implementation, the temporal gap follows a curriculum $g \in \{8,16,32,64\}$ and progressively enlarge it until reaching the full-length setting, so the model first learns short-range endpoint consistency and then gradually handles larger pose variation across time. Importantly, this pretraining formulation allows each original motion window to be converted into multiple endpoint-pair supervision samples with different temporal spans, effectively increasing the amount of usable training data by several times. As a result, the model sees much richer ego-to-pose correspondences during training, which helps alleviate the ill-posed nature of egocentric 3D human pose and motion estimation. This provides a stronger initialization for the subsequent full-sequence modeling stage.

\subsection{Human Motion Bridging Diffusion Finetuning}
%在这里，我们提出sequent bridging diffusion 模型，让motion的生成从邻近帧出发而不是从全噪声。这个motivation在于 motion数据非常的连续规整，帧之间的数值往往连续变化,直观上来说每一帧不必从0开始。 所以我们受到bridging diffusion的启发，它从给定端点开始学习与目标端点插值分布。

%在finetuning阶段，我们使用在第一阶段预训练好的模型上进行完整的human motion bridging 任务学习 from two egocentric images。去生成中间的motion片段。

%给我latex格式，首先简单介绍普通的diffusion 模型，然后介绍普通image-to-image的bridge diffusion（第二张图片）；最后详细的介绍我们的bridge模型，解决了普通bridge diffusion不直接适用motion squence的问题（总结几点为什么），然后引出我们的motion bridging diffusion model，（联合代码/nfs-stor/gaoge.han/EgoHMB-OURSX/mydiffusion/gaussian_diffusion.py和第一张图片algorithm）。

% 1. 为了适应motion这个sequeue数据，以中间为分割，每一帧参考邻近帧，这是自然的，对于两个端点自然是离的近的参考更有意义。 2.我们预测的是x_0（有助于监督几何loss，启发自mdm）且在最后让噪声趋于正态高斯（有助于以其他模态做统一的中间生成，无需一开始就依靠同模态参考）。 要讲出为什么要这样，这是我们设计的特点。 

% 我们的motion bridge diffusion model的设计

% 

% \subsection{Motion Bridging Diffusion Model}

We first briefly review standard diffusion models and conventional bridge diffusion, and then introduce our motion bridging diffusion model for human motion generation from heterogeneous conditions.

\paragraph{Standard Diffusion.}
Denoising diffusion models learn a generative process by progressively corrupting clean data with Gaussian noise and then reversing this process. Given a clean sample $x_0$, the forward process is defined as
\begin{equation}
q(x_t \mid x_0)=\mathcal{N}\!\left(x_t;\sqrt{\bar{\alpha}_t}x_0,\,(1-\bar{\alpha}_t)\mathbf{I}\right),
\end{equation}
% which can be written in reparameterized form as
% \begin{equation}
% x_t=\sqrt{\bar{\alpha}_t}x_0+\sqrt{1-\bar{\alpha}_t}\,\epsilon,\qquad \epsilon\sim\mathcal{N}(0,\mathbf{I}).
% \end{equation}
A neural network is then trained to recover the clean signal, or equivalently the injected noise, from the noisy sample $x_t$ at timestep $t$. In the standard DDPM formulation, the model predicts the noise term $\epsilon$, and generation starts from a Gaussian sample $x_T\sim\mathcal{N}(0,\mathbf{I})$ followed by iterative denoising.

\paragraph{Standard Bridge diffusion.}
Standard bridge diffusion defines a stochastic path between a source sample $x_0$ and a paired target sample $y$. Its forward process is
\begin{equation}
x_t=(1-m_t)x_0+m_t y+\sqrt{\delta_t}\,\epsilon,
\qquad \epsilon\sim\mathcal{N}(0,\mathbf{I}),
\end{equation}
and the model is trained to predict the bridge residual $m_t(y-x_0)+\sqrt{\delta_t}\epsilon$. During sampling, it starts from the conditional endpoint $x_T=y$ and iteratively recovers $x_{t-1}$ from $x_t$ using the predicted residual and the reference $y$. Therefore, standard bridge diffusion is naturally suited to paired same-modality translation, where the target-space reference $y$ is available throughout the reverse process.

\paragraph{Motivation.} Inspired by bridge diffusion, we note that its bridging idea is particularly natural for human motion. A motion sequence is inherently transitional, with each frame evolving smoothly from nearby states, so modeling motion as a stochastic bridge provides a suitable inductive bias for temporal continuity and coherent kinematics.

% Inspired by bridge diffusion, we observe that its core idea is particularly natural for human motion modeling. Human motion is inherently transitional: each frame is closely related to its temporal neighbors, and a motion sequence can be viewed as a continuous evolution between nearby states rather than a collection of independent poses. This makes the notion of constructing a stochastic bridge especially appealing for motion, as it provides a meaningful inductive bias toward temporal continuity, smooth state transitions, and coherent evolution from sparse observations to the full sequence.

\paragraph{Why Standard Bridge Diffusion is Insufficient for Motion Sequences?}
Although this formulation is natural for image-to-image translation, it is not directly suitable for human motion generation in our setting. First, the original bridge objective depends explicitly on a same-modality reference $y$ in the output space. In our task, however, the available conditions are heterogeneous, such as egocentric image observations and camera trajectory. Second, human motion is a structured temporal sequence rather than a single static sample, so the bridge should respect frame-wise temporal relations instead of treating the whole sequence as an unstructured paired object. \textbf{Solutions}: To make bridge diffusion suitable for motion sequences, we introduce two key modifications. First, instead of defining a global bridge between two fixed endpoints, we construct a local frame-wise bridge: splitting the sequence around its temporal center, each frame is paired with a nearby temporal neighbor, which is more natural for motion due to its local smoothness and short-range temporal dependency. Second, inspired by MDM, we predict the clean motion $x_0$ rather than the bridge noise, while still designing the forward process so that the terminal state approaches a standard Gaussian distribution. This allows direct supervision in motion space, which is more compatible with geometric losses, and also enables a unified generation process conditioned on heterogeneous modalities, without requiring a same-modality output reference as in standard bridge diffusion.

\paragraph{Motion Bridge Diffusion Model.}
As shown in \ref{algo:1}, we formulate bridge diffusion directly in motion space while using external modalities only as conditions. Let $X^{1:N}=\{X^1,\dots,X^N\}$ denote a motion sequence, where $X^i$ is the motion representation at frame $i$. We additionally define zero-padded boundaries $X^0=X^{N+1}=0$. For each frame $i$, we construct a local bridge pair
\begin{equation}
x_0^i=X^i,\qquad y^i=X^{i+s_i},
\end{equation}
where
\begin{equation}
s_i = 2\cdot \mathbb{I}_{i>N/2}-1.
\end{equation}
This means that frames in the first half of the sequence are bridged toward their previous neighbors, while frames in the second half are bridged toward their next neighbors. The boundary terms are naturally handled by the zero padding. In this way, the bridge is defined between temporally adjacent motion states rather than between two global samples, which makes it more suitable for modeling motion dynamics.

% At the same time, we associate the motion sequence with sparse egocentric observations $(o^1,o^N)$ corresponding to the first and last frames $(X^1,X^N)$. These observations are not used as bridge endpoints in motion space; instead, they serve as conditioning signals that guide generation.

For the forward process, based on this local bridge construction, we define the forward noising process as
\begin{equation}
x_t^i
=
\sqrt{\bar{\alpha}_t}\left((1-m_t)x_0^i+m_t y^i\right)
+
\sqrt{1-\bar{\alpha}_t}\,\epsilon^i,
\qquad
\epsilon^i\sim\mathcal{N}(0,\mathbf{I}),
\end{equation}
where $m_t\in[0,1]$ is the bridge coefficient and $\bar{\alpha}_t$ follows the standard diffusion noise schedule. In our implementation, $m_t$ increases linearly with $t$. This process can be interpreted as first defining a motion-dependent bridge mean between the current frame and its local temporal reference, and then applying standard Gaussian diffusion around this mean.

\begin{wrapfigure}{r}{0.65\linewidth}
\vspace{-10pt}
\begin{minipage}{\linewidth}

\begin{algorithm}[H]
\caption{Motion Bridging Diffusion Model Training}
\label{algo:1}

\Repeat{converged}{
    Sample motion sequence data $X^{1:N}\sim q(X^{1:N})$, with $X^{0}=X^{N+1}=0$\;
    Sample a pair of egocentric images $(o^{1},o^{N})\sim q(O^{1},O^{N})$\;
    Sample timestep $t\sim \mathrm{Uniform}(1,\ldots,T)$\;

    \For{$i=1,\ldots,N$}{
        Define $s_i = 2\cdot \mathbbm{1}_{i>N/2}-1$\;
        Define paired data $(x_0^i,y^i)=(X^i,X^{i+s_i})$\;
        Sample Gaussian noise $\varepsilon^i\sim \mathcal{N}(0,I)$\;
        Define condition
        $
            c^i=\mathbbm{1}_{i=1}\cdot o^1
            +\mathbbm{1}_{i=N}\cdot o^N
            +0
        $\;
        Forward diffusion:
        $
            x_t^i
            =
            \sqrt{\bar{\alpha}_t}
            \bigl((1-m_t)x_0^i+m_t y^i\bigr)
            +
            \sqrt{1-\bar{\alpha}_t}\,\varepsilon^i
        $\;
    }

    Take a gradient descent step on:
    $
        \nabla_{\theta}
        \sum_{i=1}^{N}
        \left\|
            x_0^i
            -
            \epsilon_{\theta}(x_t^i,t,c^i)
        \right\|^2
    $\;
}

\end{algorithm}
\end{minipage}
\vspace{-10pt}
\end{wrapfigure}

This design is important because it preserves a key property of conventional diffusion models: as $t$ approaches the final timestep $T$, $\bar{\alpha}_t\rightarrow 0$, and thus
\begin{equation}
x_T^i \approx \epsilon^i \sim \mathcal{N}(0,\mathbf{I}).
\end{equation}
% although the intermediate states follow a motion-aware bridge, the terminal distribution still approaches a standard Gaussian.
Therefore, this allows sampling to start from a modality-agnostic Gaussian prior, which is crucial for unified generation from heterogeneous input instead of a human pose.

During sampling, the process starts from Gaussian noise. At each timestep $t$, the model predicts the clean motion $\hat{x}_0 = \epsilon_\theta(x_t,t,c)$ and uses it to compute the reverse transition to $x_{t-1}$. Repeating this process from $t=T$ to $t=1$ gradually removes noise and yields the final motion sample $x_0$.

\section{Experiments}

In this section, we present in detail the strong baselines constructed for the new egocentric human motion bridging task, as well as the technical implementation, evaluation metrics, and dataset details. These experimental designs are intended to demonstrate the unified formulation and effectiveness of our method on this task.
% 在这个章节，我们将详细介绍我们如何对egocentric human motion bridging这个新任务构建比较的strong baseline以及技术实现、指标、数据集详情。从而证实我们方法对这个任务的统一性和有效性。

\subsection{Implementation}

% We adopt a conditional Gaussian diffusion model to model the entire human motion sequence. Given a motion representation sequence $\mathbf{x}_0$ of length $T = 80$, during training we first add noise to $\mathbf{x}_0$ at a randomly sampled diffusion step $t$ to obtain $\mathbf{x}_t$, and then use a conditional denoising network to predict the clean sequence $\mathbf{x}_0$. Specifically, the trajectory features are provided as frame-wise vectors, which are mapped into the latent space through a linear layer and then directly added to the motion tokens. 

We adopt a conditional bridge diffusion model to model the entire human motion sequence. The denoising network is based on a Transformer decoder architecture with 12 layers, 12 attention heads, and a latent dimension of 768. The input to the network consists of noisy motion tokens, while the conditioning information includes egocentric image features. The image features, together with the diffusion timestep embedding, form context tokens that are injected into the denoising process through cross-attention. 

For the egocentric human motion bridging task, we retain only the conditioning information from the first and last frames, while masking out all trajectory and image conditions for the intermediate frames. This forces the model to recover the complete intermediate motion based solely on the endpoint constraints. The training objective is to predict the clean motion representation itself, and an MSE loss is computed over the valid frames. We adopt a head-centric motion representation, rather than the pelvis-centric representation commonly used in motion synthesis literature, which makes it better suited to egocentric devices. We use DINOv2~\cite{oquab2023dinov2} fine-grained features to extract representations from egocentric images, since they have been shown to be more effective than CLIP in~\cite{Patel_2025_ICCV}.

\subsection{Dataset}
% 我们使用了large-scale EgoExo4D~\cite{grauman2024egoexo4d} dataset带有SMPL-X~\cite{bogo2016smplify}标注的版本数据集EE4D-Motion~\cite{Patel_2025_ICCV」 dataset
 We use EE4D-Motion~\cite{Patel_2025_ICCV}, a version of the large-scale EgoExo4D~\cite{grauman2024egoexo4d} dataset annotated with SMPL-X~\cite{bogo2016smplify}. It is a 3D human motion dataset comprising 208 hours of time-synchronized 3D motion data, egocentric videos, and associated annotations. The dataset captures real-world human activities performed by participants wearing Project Aria glasses, covering diverse scenarios such as cooking, dance, sports, music, healthcare, and bike repair. To construct the training set, we extract 8-second video clips sampled at 10 fps ($\mathrm{N}=80$) with a sliding window of 2 seconds, resulting in 143K training samples. For validation, we use the same clip length and frame rate but sample clips every 20 seconds, yielding 4,400 validation samples. 
 % The egocentric videos are paired with the 3D trajectory of the ego camera, and the dataset further provides SMPL-X body model representations for 3D human motion modeling and activity understanding.

%  \begin{table}[t]
% \centering
% \footnotesize
% \caption{\textbf{Model Efficiency Comparison}: Comparison of model complexity and inference latency on an NVIDIA A100 GPU.}
% \label{tab:efficiency_compare}
% \begin{tabular}{p{3.2cm}cc}
% \toprule
% \thead{Method} & \thead{Params (M)} & \thead{Latency (ms)$\downarrow$} \\
% \cmidrule{1-3}
% MDM~\cite{tevet2022human} &  &  \\
% UEM~\cite{li2023egoego} &  &  \\
% EgoAllo~\cite{yi2024egoallo} + UEM~\cite{li2023egoego} &  &  \\
% Ours & \textbf{} & \textbf{} \\
% \bottomrule
% \end{tabular}
% \end{table}

\begin{wraptable}{r}{0.42\textwidth}
% \vspace{-0.5em}
\vspace{-10pt}
\centering
\footnotesize
\caption{\textbf{Model Efficiency Comparison}: Comparison of model complexity and inference latency for generating a 1-second clip using 1000 diffusion steps on an NVIDIA A100 GPU.}
\label{tab:efficiency_compare}
\setlength{\tabcolsep}{4pt}
\begin{tabular}{lcc}
\toprule
Method & Params & Latency $\downarrow$ \\
\midrule
MDM~\cite{tevet2022human} & 87M & 1.66s \\
UEM~\cite{li2023egoego} & 89M & 1.82s \\
EgoAllo~\cite{li2023egoego}+UEM & 140M & 3.21s \\
Ours & \textbf{86M} & \textbf{1.58s} \\
\bottomrule
\end{tabular}
\vspace{-10pt}
\end{wraptable}

\subsection{Metrics}
%egocentric human motion bridging task需要在首尾端点处pose与GT尽量对齐的情况下去生成中间的motion。因此我们的评估也分为对端点帧使用estimation指标和对中间motion使用生成指标评估. 
In the egocentric human motion bridging task, the goal is to generate the intermediate motion while ensuring that the poses at the start and end points are as aligned as possible with the ground truth. Accordingly, our evaluation is divided into two parts: estimation metrics for the endpoint frames and generation metrics for the intermediate motion.

We evaluate endpoint pose estimation quality using a diverse set of metrics in the 3D geometric space. \textbf{MPJPE} measures the mean per-joint position error (in meters) over 22 body joints. \textbf{MPJPE-PA} further aligns the predicted and ground-truth motions on a per-frame basis using Procrustes analysis before computing MPJPE, thereby reflecting the accuracy of local pose estimation. \textbf{MPJPE-B} reports the mean per-joint position error (in meters) for the body joints.  \textbf{MPJPE-H} reports the mean per-joint position error (in meters) for the hand joints. 

% We use \textbf{Head Rotation Error} and \textbf{Head Translation Error} to assess head consistency, which quantify the rotational discrepancy and translational error (in meters) of the head joint, respectively. 

% These metrics are particularly important for evaluating how well the model satisfies head-aligned conditioning signals in the estimation setting. Following prior work, the rotation error is computed as the Frobenius norm of the relative rotation matrix difference~\cite{48}.

For the motion generation part, we consider contact-related motion quality metrics. \textbf{Foot Sliding}~\cite{han2025reindiffuse} measures the amount of undesired foot motion when the foot is close to the ground. \textbf{Foot Contact}~\cite{ han2025reindiffuse} computes the average distance (in meters) between the foot and the ground, capturing both floating artifacts and ground penetration. In addition, we adopt semantic-level motion evaluation metrics. \textbf{Semantic Similarity} measures the consistency between generated motions and ground-truth motions in a semantic latent space, analogous to CLIP-score~\cite{hessel2021clipscore}. Concretely, we use the motion encoder from TMR~\cite{guo2022tm2t} to project motions into the latent space and compute the cosine distance between their embeddings. \textbf{FID} measures the distributional gap between generated and real motions in the same latent space, similar to the standard FID used in image generation. \textbf{Motion Jitter} measures the average variation in motion within a fixed-length temporal window; lower values indicate smoother motion.

\subsection{Baselines}

Since there are no direct baselines for the egocentric human motion bridging task, we construct strong baselines based on state-of-the-art motion modeling paradigms. Specifically, we compare our method against two categories of strong baselines: one-stage methods (MDM~\cite{tevet2022human}, UniEgoMotion (UEM)~\cite{Patel_2025_ICCV}, Genmo~\cite{li2025genmo}) and two-stage methods (EgoAllo~\cite{yi2024egoallo} with UniEgoMotion~\cite{Patel_2025_ICCV}). 

% MoMask~\cite{guo2024momask}

One-stage methods directly model motion generation in a conditional manner, whereas two-stage methods first estimate the 3D poses at the start and end frames and then feed them into an in-betweening model to predict the intermediate motion sequence. For a fair comparison, we retrain all baseline methods on the EE4D-Motion~\cite{Patel_2025_ICCV,grauman2024egoexo4d} dataset and keep the same input and output conditions to adapt them to our proposed human motion bridging task.

%Since there are no direct baselines for egocentric human motion bridging task, we construct strong baselines based on state-of-the-art motion modeling practices. 我们比较我们的方法与两类strong baseline，分别是一阶段方法（Uniegomotion， EgoEgo）和二阶段方法（MDM + wham）。一阶段是指以条件建模的生成模型方式；二阶段是指先使用estimation的方式估计出首尾的3D pose再送入in- between模型估计出中间运动 为了公平比较，我们将它们的方法 在EE4D-Motion dataset重新训练，保持同样的输入和输出条件以适应 our proposed human motion bridging task。

\begin{figure}
    \centering
    \includegraphics[width=0.99\linewidth]{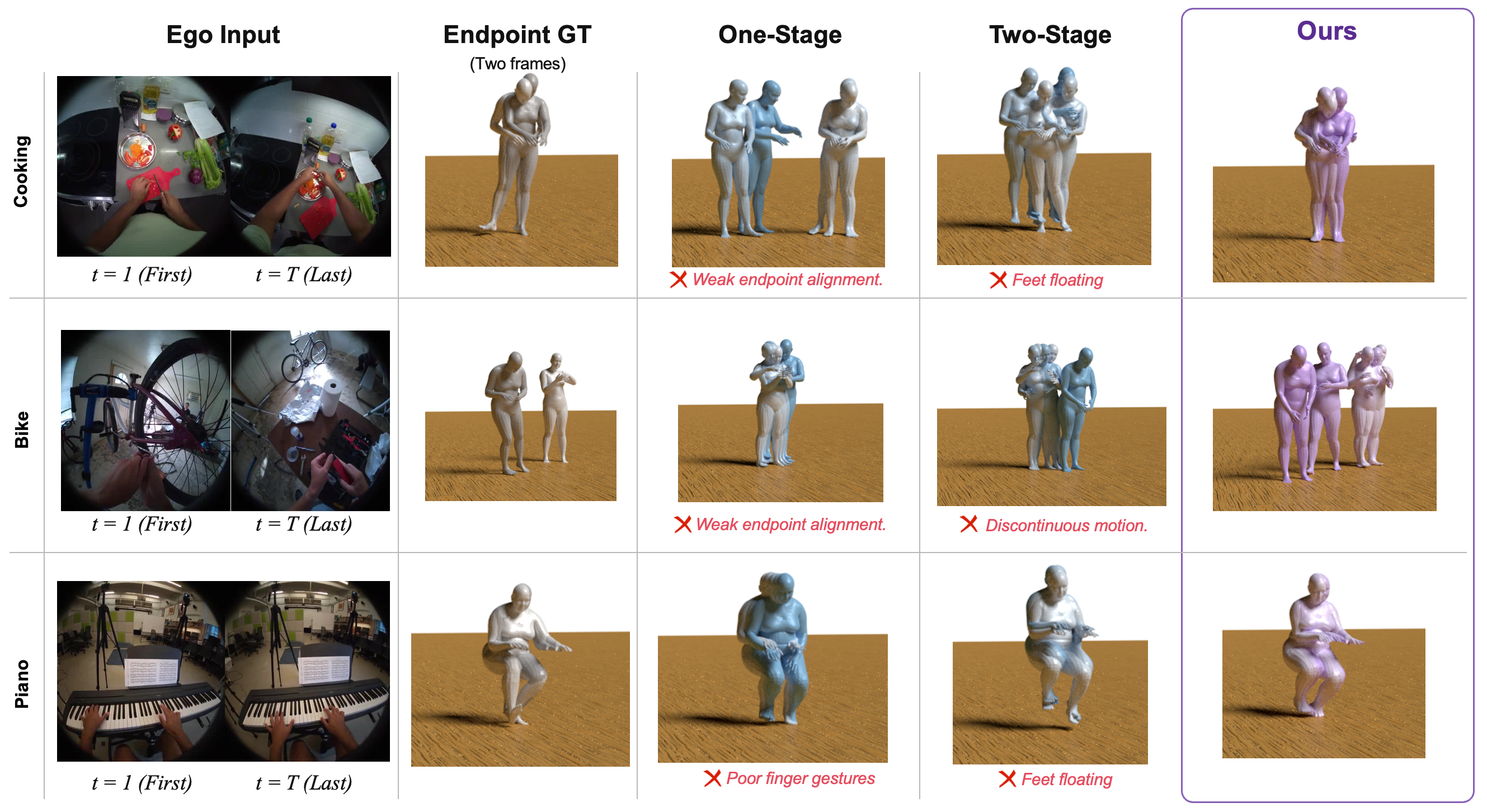}
    \caption{\textbf{Qualitative Comparison with Strong Baselines.} The first two columns show the input start (bright white) / end (dark gray) egocentric images and the corresponding ground-truth endpoint poses. The next two columns present the one-stage baseline (UEM) and the two-stage baseline (EgoAllo+UEM), respectively, followed by our method in the last column. Further comparisons demonstrating physical continuity are available in the \textbf{supplementary material video}.}
    \label{fig:ego_vis}
\end{figure}

\begin{table*}\centering 
\footnotesize
\caption{\textbf{Endpoint Results}: Comparison with strong baselines for endpoint pose estimation.}

\begin{tabular}{lcccc}
\toprule
\thead{Method} & \thead{MPJPE($\downarrow$)} & \thead{MPJPE-PA($\downarrow$)} & \thead{MPJPE-B($\downarrow$)} & \thead{MPJPE-H($\downarrow$)} \\
\cmidrule{1-5}
MDM~\cite{tevet2022human} & 0.742 & 0.286 & 0.701 & 0.865 \\
Genmo~\cite{li2025genmo} & 0.524 & 0.213 & 0.537 & 0.646 \\
UEM~\cite{Patel_2025_ICCV} & 0.486 & 0.162 & 0.463 & 0.579 \\
EgoAllo~\cite{yi2024egoallo} + UEM & 0.421 & 0.137 & 0.398 & 0.512 \\
Ours & $\mathbf{0.327}$ & $\mathbf{0.091}$ & $\mathbf{0.309}$ & $\mathbf{0.409}$ \\
\bottomrule
\end{tabular}
\vspace{-10pt}
\label{tab:ego_endpoint}
\end{table*}

\begin{table*}\centering
\footnotesize
\caption{\textbf{Motion Generation Results}: Comparison with strong baselines on EE4D-Motion for motion generation quality and diversity.}
\begin{tabular}{lcccccc}
\toprule
\thead{Method} & \thead{Foot\\Slide($\downarrow$)} & \thead{Foot\\Contact($\downarrow$)} & \thead{Semantic\\Sim.($\uparrow$)} & \thead{FID($\downarrow$)} & \thead{Jitter($\downarrow$)} & \thead{Diversity} \\
\cmidrule{1-7}
MDM~\cite{tevet2022human} & 7.318 & 0.087 & 0.049 & 0.132 & 0.419 & 9.325 \\
Genmo~\cite{li2025genmo} & 5.210 & 0.053 & 0.076 & 0.087 & 0.378 & 9.274 \\
UEM~\cite{Patel_2025_ICCV} & 4.912 & 0.049 & 0.074 & 0.071 & 0.352 & 9.457 \\
EgoAllo~\cite{yi2024egoallo} + UEM & 4.865 & 0.047 & 0.075 & 0.069 & 0.347 & 9.411 \\
Ours & $\mathbf{3.446}$ & $\mathbf{0.031}$ & $\mathbf{0.082}$ & $\mathbf{0.047}$ & $\mathbf{0.176}$ & 9.403 \\
\bottomrule
\end{tabular}
\vspace{-10pt}
\label{tab:ego_generation}
\end{table*}

\begin{table*}\centering
\footnotesize
\setlength{\tabcolsep}{4pt}
\caption{\textbf{Cross-dataset Generalization on Nymeria~\cite{ma2024nymeria}.} Models trained on EE4D-Motion are evaluated directly on Nymeria without training or fine-tuning on Nymeria.}
\label{tab:nymeria}
\begin{tabular}{lccccccc}
\toprule
\thead{Method} & \thead{Endpoint\\MPJPE($\downarrow$)} & \thead{Foot\\Slide($\downarrow$)} & \thead{Foot\\Contact($\downarrow$)} & \thead{Semantic\\Sim.($\uparrow$)} & \thead{FID($\downarrow$)} & \thead{Jitter($\downarrow$)} & \thead{Diversity} \\
\midrule
MDM~\cite{tevet2022human} & 0.875 & 6.452 & 0.104 & 0.061 & 0.164 & 0.537 & 9.347 \\
Genmo~\cite{li2025genmo} & 0.643 & 4.895 & 0.081 & 0.071 & 0.126 & 0.418 & 9.445 \\
UEM~\cite{Patel_2025_ICCV} & 0.530 & 3.875 & 0.062 & 0.072 & 0.114 & 0.352 & 9.576 \\
EgoAllo~\cite{yi2024egoallo} + UEM & 0.481 & 3.676 & 0.053 & 0.076 & 0.093 & 0.324 & 9.501 \\
Ours & $\mathbf{0.369}$ & $\mathbf{2.748}$ & $\mathbf{0.023}$ & $\mathbf{0.086}$ & $\mathbf{0.059}$ & $\mathbf{0.153}$ & 9.533 \\
\bottomrule
\end{tabular}
\vspace{-10pt}
\end{table*}

\subsection{Results}

\noindent \textbf{Quantitative Results:} Tables~\ref{tab:ego_endpoint} and~\ref{tab:ego_generation} compare our method with one-stage and two-stage baselines on endpoint estimation and motion generation, respectively. MDM, Genmo, and UEM directly generate motion in a one-stage manner, while EgoAllo+UEM adopts a two-stage pipeline that first estimates endpoint poses and then performs motion generation. Our conditional unified motion bridge diffusion model achieves the best endpoint accuracy and motion quality scores while maintaining comparable Diversity (9.403). In particular, the lower endpoint MPJPE-based errors indicate more accurate endpoint pose estimation, while the improved Foot Slide, Foot Contact, Semantic Similarity, FID, and Jitter scores demonstrate more plausible, temporally stable motion generation. Besides, Table~\ref{tab:efficiency_compare} demonstrates that our unified bridge diffusion design achieves better model inference performance compared with other models.

We further evaluate cross-dataset generalization by directly applying models trained on EE4D-Motion to Nymeria, without training or fine-tuning on the target dataset (Table~\ref{tab:nymeria}). Our method achieves the lowest endpoint MPJPE (0.369), FID (0.059), and Jitter (0.153), together with the best foot-related metrics and Semantic Similarity. Diversity remains comparable to the baselines at 9.533. These zero-shot results support the transfer of our endpoint estimation and motion generation capabilities beyond the training dataset.

\noindent \textbf{Qualitative Results:} Figure~\ref{fig:ego_vis} compares motion sequences generated from the same start and end egocentric images. The one-stage baseline exhibits weak endpoint alignment and implausible hand configurations, while the two-stage baseline improves the global structure but still produces physical artifacts such as floating feet. Our method better aligns the endpoints with the reference poses and preserves more plausible body configurations and contact patterns throughout the sequence. These observations agree with the quantitative results in Tables~\ref{tab:ego_endpoint} and~\ref{tab:ego_generation}, supporting the benefit of jointly modeling endpoint estimation and motion bridging. Additional comparisons of temporal continuity are provided in the supplementary video.

\subsection{Ablation Study}
Table~\ref{tab:ablation} presents a controlled $2\times2$ comparison of standard diffusion and bridge diffusion, each with and without E2C endpoint pretraining. All variants use the same architecture and experimental protocol, allowing us to examine the two components separately.

\begin{figure}[!htbp]
    \centering
    \includegraphics[width=0.68\linewidth]{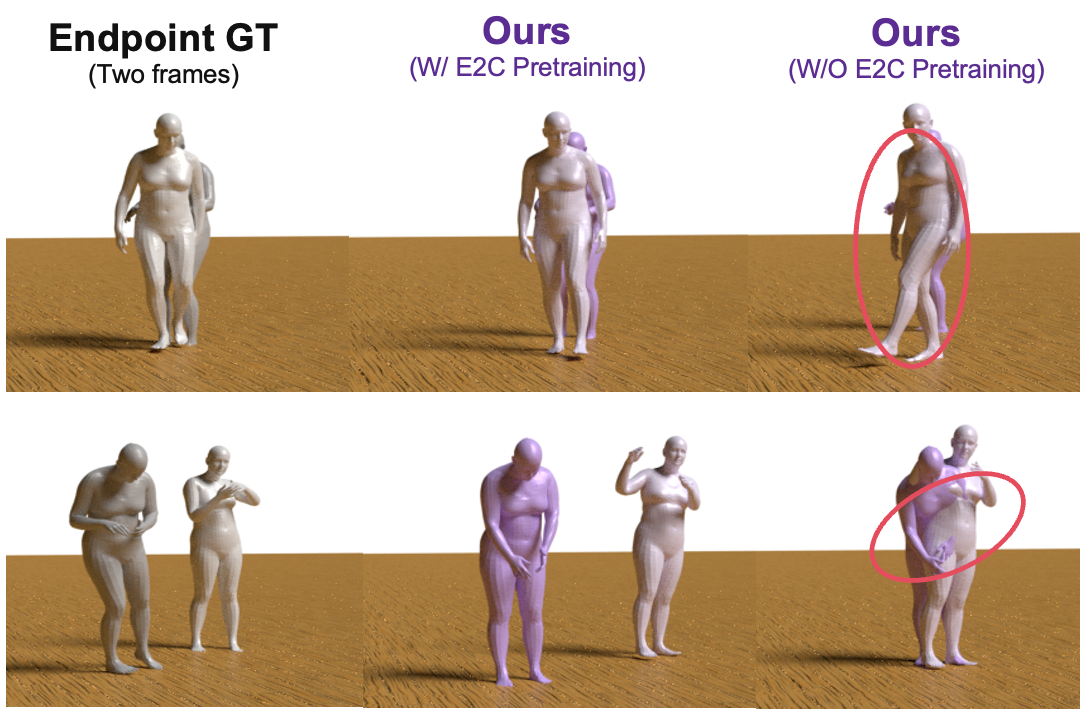}
    \caption{Visual Comparison of E2C Pretraining Ablation.}
    \label{fig:abla-pre}
\end{figure}

\noindent \textbf{Effect of E2C Pretraining.} E2C reduces endpoint MPJPE from 0.468 to 0.362 for standard diffusion and from 0.431 to 0.327 for bridge diffusion. Its benefit therefore holds under both diffusion formulations. Figure~\ref{fig:abla-pre} provides complementary visual evidence: compared with the variant without E2C, the pretrained model produces endpoint body configurations that more closely match the ground truth. The highlighted discrepancies without pretraining illustrate the ambiguity of recovering body poses from egocentric views, in which much of the body is unobserved. E2C also improves all reported motion generation metrics under both diffusion formulations.

\noindent \textbf{Effect of Bridge Diffusion.} With E2C fixed, replacing standard diffusion with bridge diffusion reduces Foot Slide from 4.847 to 3.446, FID from 0.094 to 0.047, and Jitter from 0.308 to 0.176, while increasing Semantic Similarity from 0.057 to 0.082. The same trend holds without E2C, indicating that the generation improvements are not solely attributable to endpoint pretraining. Together, Figure~\ref{fig:abla-pre} and Table~\ref{tab:ablation} show complementary benefits: E2C strengthens endpoint alignment, while bridge diffusion improves motion quality and temporal smoothness. Combining both yields the best results across all metrics reported in the ablation.

\begin{table*}\centering 
\footnotesize
\caption{\textbf{Ablation Study.} A controlled comparison of standard diffusion and bridge diffusion with and without E2C endpoint pretraining. Endpoint MPJPE evaluates endpoint-pose accuracy; the remaining metrics evaluate motion generation quality.}

\begin{tabular}{lcccccc}
\toprule
\multicolumn{1}{c}{\thead{Method}} & \thead{Endpoint\\MPJPE($\downarrow$)} & \thead{Foot\\Slide($\downarrow$)} & \thead{Foot\\Contact($\downarrow$)} & \thead{Semantic\\Sim.($\uparrow$)} & \thead{FID($\downarrow$)} & \thead{Jitter($\downarrow$)} \\
\cmidrule{1-7}
Standard Diffusion w/o E2C & $0.468$ & $4.921$ & $0.058$ & $0.052$ & $0.097$ & $0.325$ \\
Bridge Diffusion w/o E2C & $0.431$ & $3.684$ & $0.038$ & $0.074$ & $0.056$ & $0.184$ \\
Standard Diffusion w/ E2C & $0.362$ & $4.847$ & $0.056$ & $0.057$ & $0.094$ & $0.308$ \\
Bridge Diffusion w/ E2C & $\mathbf{0.327}$ & $\mathbf{3.446}$ & $\mathbf{0.031}$ & $\mathbf{0.082}$ & $\mathbf{0.047}$ & $\mathbf{0.176}$ \\
\bottomrule
\end{tabular}
\vspace{-10pt}
\label{tab:ablation}
\end{table*}

\section{Conclusion and Limitation}

In this paper, we introduced Egocentric Human Motion Bridging, a new task that generates coherent 3D human motion segments from sparse egocentric endpoint observations. To address the challenges of partial body visibility, endpoint ambiguity, and temporal ill-posedness, we proposed Ego-HMB, a unified bridge diffusion framework that jointly internalizes endpoint pose estimation and temporally consistent motion completion through a two-stage training strategy. Extensive experiments demonstrate that our method outperforms both direct conditional generation and two-stage estimation--interpolation baselines, producing more accurate, continuous, and plausible motions. A current limitation of this work is that we focus on human motion bridging from egocentric images, which are commonly available in VR/AR and robotics scenarios, while other modalities such as text and audio are not yet explored. Nevertheless, our framework can be naturally extended to incorporate these additional modalities as conditioning signals. Another promising direction is to further adapt the generated motion representations for direct humanoid robot motion control, enabling broader embodied applications.

\begin{ack}
Mingming Gong was supported by ARC DP240102088 and WIS--MBZUAI 142571.
% Use unnumbered first level headings for the acknowledgments. All acknowledgments

% Do {\bf not} include this section in the anonymized submission, only in the final paper. You can use the \texttt{ack} environment provided in the style file to automatically hide this section in the anonymized submission.
\end{ack}

{
    \small
    \bibliographystyle{plain} % 或 ieeetr、acm 等
    \bibliography{neurips_2026.bib}
}

%%%%%%%%%%%%%%%%%%%%%%%%%%%%%%%%%%%%%%%%%%%%%%%%%%%%%%%%%%%%

% \appendix

% \section{Technical appendices and supplementary material}
% Technical appendices with additional results, figures, graphs, and proofs may be submitted with the paper submission before the full submission deadline (see above). You can upload a ZIP file for videos or code, but do not upload a separate PDF file for the appendix. There is no page limit for the technical appendices. 

% Note: Think of the appendix as ``optional reading'' for reviewers. The paper must be able to stand alone without the appendix; for example, adding critical experiments that support the main claims to an appendix is inappropriate. 

%%%%%%%%%%%%%%%%%%%%%%%%%%%%%%%%%%%%%%%%%%%%%%%%%%%%%%%%%%%%

\end{document}